\documentclass[sigconf]{acmart}
\AtBeginDocument{%
  }
\copyrightyear{2025}
\acmYear{2025}
\setcopyright{acmlicensed}
\acmConference[CIKM '25] {Proceedings of the 34th ACM International Conference on Information and Knowledge Management}{ November 10--14, 2025}{Seoul, Republic of Korea.}
\acmBooktitle{Proceedings of the 34th ACM International Conference on Information and Knowledge Management (CIKM '25), November 10--14, 2025, Seoul, Republic of Korea}
\acmISBN{979-8-4007-2040-6/2025/11}
\acmDOI{10.1145/3746252.3761056}
\usepackage{framed}
\usepackage{balance}
\usepackage{float}
\usepackage{hyperref}
\usepackage{subcaption}
\usepackage{graphicx}
\usepackage{placeins}
\usepackage{amsmath} 
\usepackage{multirow}
\usepackage{url}
\usepackage{caption}
\usepackage{dblfloatfix}
\usepackage{enumitem}
\begin{document}
\title[TableTime: A Table Understanding Paradigm for Time Series Classification]{TableTime: Reformulating Time Series Classification as Training-Free Table Understanding with Large Language Models}
\author{Jiahao Wang}
\affiliation{%
  \institution{State Key Laboratory of Cognitive Intelligence,  University of Science and Technology of China}
  \city{Hefei}
  \state{Anhui Province}
  \country{China}
}
\email{SA24229078@mail.ustc.edu.cn}

\author{Mingyue Cheng}
\authornote{Mingyue Cheng is the corresponding author.}
\affiliation{%
  \institution{State Key Laboratory of Cognitive Intelligence,  University of Science and Technology of China}
  \city{Hefei}
  \state{Anhui Province}
  \country{China}
}
\email{mycheng@ustc.edu.cn}

\author{Qingyang Mao}
\affiliation{%
  \institution{State Key Laboratory of Cognitive Intelligence,  University of Science and Technology of China}
  \city{Hefei}
  \state{Anhui Province}
  \country{China}
}
\email{maoqy0503@mail.ustc.edu.cn}

\author{Yitong Zhou}
\affiliation{%
  \institution{State Key Laboratory of Cognitive Intelligence,  University of Science and Technology of China}
  \city{Hefei}
  \state{Anhui Province}
  \country{China}
}
\email{yitong.zhou@mail.ustc.edu.cn}

\author{Daoyu Wang}
\affiliation{%
  \institution{State Key Laboratory of Cognitive Intelligence,  University of Science and Technology of China}
  \city{Hefei}
  \state{Anhui Province}
  \country{China}
}
\email{wdy030428@mail.ustc.edu.cn}

\author{Qi Liu}
\affiliation{%
  \institution{State Key Laboratory of Cognitive Intelligence,  University of Science and Technology of China}
  \city{Hefei}
  \state{Anhui Province}
  \country{China}
}
\email{qiliuql@ustc.edu.cn}

\author{Feiyang	Xu}
\affiliation{
  \institution{Artificial Intelligence Research Institute, iFLYTEK Co., Ltd}
  \city{Hefei}
  \state{Anhui Province}
  \country{China}
}
\email{fyxu2@iflytek.com}

\author{Xin Li}
\affiliation{%
  \institution{State Key Laboratory of Cognitive Intelligence,  University of Science and Technology of China}
  \city{Hefei}
  \state{Anhui Province}
  \country{China}
}
\email{leexin@ustc.edu.cn}
\renewcommand{\shortauthors}{Jiahao Wang et al.}

\begin{abstract}
Large language models (LLMs) have shown promise in multivariate time series classification (MTSC). To effectively adapt LLMs for MTSC, it is crucial to generate comprehensive and informative data representations. Most methods utilizing LLMs encode numerical time series into the model’s latent space, aiming to align with the semantic space of LLMs for more effective learning.
Despite effectiveness, we highlight three limitations that these methods overlook: (1) they struggle to incorporate temporal and channel-specific information, both of which are essential components of multivariate time series; (2) aligning the learned representation space with the semantic space of the LLMs proves to be a significant challenge; (3) they often require task-specific retraining, preventing training-free inference despite the generalization capabilities of LLMs. To bridge these gaps, we propose TableTime, which reformulates MTSC as a table understanding task. Specifically, TableTime introduces the following strategies: (1) utilizing tabular form to unify the format of time series, facilitating the transition from the model-centric approach to the data-centric approach; (2) representing time series in text format to facilitate seamless alignment with the semantic space of LLMs; (3) designing a knowledge-task dual-driven reasoning framework, TableTime, integrating contextual information and expert-level reasoning guidance to enhance LLMs' reasoning capabilities and enable training-free classification.
Extensive experiments conducted on 10 publicly available benchmark datasets from the UEA archive validate the substantial potential of TableTime to be a new paradigm for MTSC. The code is publicly available\footnote{\url{https://github.com/realwangjiahao/TableTime}}.
\end{abstract}



\maketitle
\section{Introduction}
Multivariate time series~\cite{time1,time2} are commonly encountered in various domains, consisting of sequences of events collected over time, where each event includes observations recorded across multiple attributes.
For example, the electrocardiogram (ECG)~\cite{sarkar2020self} signals in electronic health records (EHRs) capture various aspects of heart function through multiple sensors. Comprehensive analysis of such data can facilitate decision-making in real applications~\cite{use1}, such as human activity recognition, healthcare monitoring, and industry detection. Particularly, as a fundamental problem in time series analysis, multivariate time series classification (MTSC) has attracted significant attention both in academia and industry~\cite{ismail2019deep}.

\begin{figure}[t]
    \centering
    \includegraphics[width=0.48\textwidth]{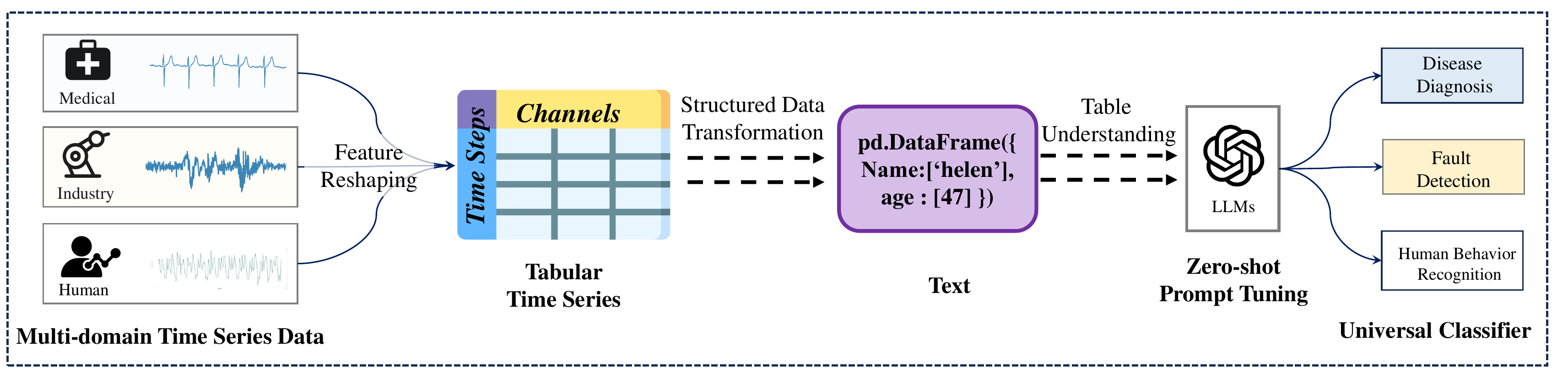}
    \caption{Illustrating the core idea of TableTime: transforming time series into tabular representations for universal classification in a data-centric paradigm.}
    \Description{Illustrating the core idea of TableTime: transforming time series into tabular representations for universal classification in a data-centric paradigm.}
    \label{motivation}
\end{figure}

Over the years, various approaches have been developed to address the MTSC task \cite{susto2018time,ismail2019deep,middlehurst2024aeon}. Traditional methods like Dynamic Time Warping (DTW) \cite{DTW} with nearest neighbor classifiers~\cite{KNN} align time series of varying lengths but struggle to capture hidden features, leading to inaccurate modeling. Machine learning methods, such as Support Vector Machines \cite{svm} and Random Forests \cite{random}, rely on handcrafted features and assume stationarity, which hinders their effectiveness in handling dynamic and diverse time series data. Deep learning \cite{deep1} has since reduced the need for feature engineering, with CNN-based \cite{cnn1,cnn2} and transformer-based \cite{informer,autoformer} approaches gaining significant attention. However, deep learning methods rely on a large amount of labeled data, limiting their application in practical scenarios. 

Recent studies have demonstrated that large language models (LLMs) exhibit robust pattern recognition and reasoning abilities over complex sequences \cite{minaee2024large}, which has spurred growing interest in their application to time series analysis \cite{gruver2024large, tao2024hierarchical}. LLM-based methods for time series analysis can generally be classified into two categories: prompt-based methods and retraining-based methods. Prompt-based methods, exemplified by PromptCast~\cite{xue2023promptcast,liu2024lstprompt}, directly apply LLMs to downstream tasks by constructing tasks in a sentence-to-sentence format. In contrast, retraining-based methods involve modifying some or all parameters of the LLMs to adapt them to specific tasks \cite{instruct, timellm}. Both approaches enable LLMs to leverage their advanced pattern recognition and reasoning capabilities, utilizing pre-trained knowledge to capture temporal dependencies and make more accurate and generalizable predictions.

While LLM-based methods have demonstrated effectiveness, several bottlenecks remain when applying them to time series classification~\cite{LLM}. First, a mismatch exists between numerical time series and the textual semantic space of LLMs, which restricts their capacity to process numeric data effectively. Second, they struggle to capture temporal dynamics and channel-specific features, which are crucial for accurate modeling. Third, extensive fine-tuning incurs high computational costs, particularly in resource-intensive applications. Lastly, they struggle to fully release the reasoning capabilities of LLMs. These challenges highlight the need for more efficient, generalizable, and domain-adaptive approaches for MTSC.

An effective LLM-based method for MTSC, in our view, should possess several key attributes. First, it should align the numerical time series with the textual semantic space of LLMs, since LLMs are designed to process text-based inputs. Second, it should be capable of extracting both temporal consistency and inter-channel features from the time series data. Third, the method should enable training-free classification, leveraging the world knowledge acquired during the pre-training phase of the LLM. Finally, LLM-based models should possess strong reasoning abilities to handle complex tasks that require understanding both logical relationships and intricate dependencies within the data. 

To this end, we propose TableTime, a training-free classification framework based on table understanding for MTSC task. We convert the numeric time series into tabular format, preserving both temporal consistency and channel-specific information. To align the tabular time series with the semantic space of LLMs, we introduce table encoding, which converts the tabular time series into a textual representation. For training-free classification, we adopt a table understanding approach, which reformulates the MTSC task in a way that enables LLMs to classify without task-specific retraining. To maximize the reasoning potential of the LLMs, we develop a prompt incorporating neighbor-assisted enhancement and multi-path reasoning, aiming to fully leverage the LLMs' reasoning capabilities. As shown in Figure \ref{motivation}, TableTime provides a new paradigm for MTSC. To summarize, our contributions include:

\begin{itemize}[leftmargin=*]
\item We propose the table understanding paradigm for MTSC and provide detailed explanations of how it helps alleviate the bottlenecks of most existing methods.
\item Under our proposed paradigm, we design a training-free framework called TableTime, which aims to leverage the reasoning capability of LLMs for time series classification.
\item We conduct comprehensive experiments on ten benchmark multivariate time series datasets, validating the effectiveness of table understanding paradigm and TableTime framework.
\end{itemize}

\section{Related Work}
\subsection{Time Series Classification}
\begin{figure*}[ht]
    \centering
    \includegraphics[width=\textwidth]{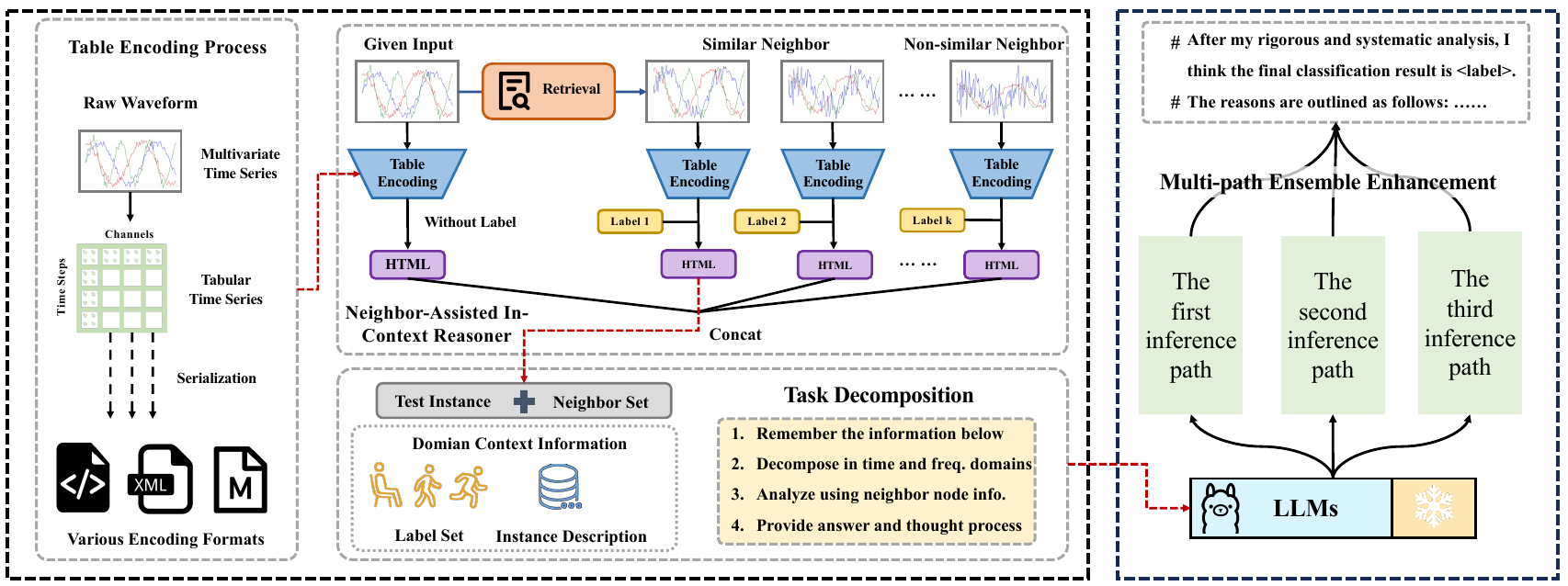}
    \caption{Illustration of the TableTime, i.e., a paradigm for MTSC based on table understanding. (Left): TableTime implements classification through neighbor retrieval and context information modeling. (Right): Multi-path ensemble enhancement.}
    \Description{Illustration of the TableTime, i.e., a paradigm for MTSC based on table understanding. (Left): TableTime implements classification through neighbor retrieval and context information modeling. (Right): Multi-path ensemble enhancement.}
    \label{main_figure}
\end{figure*}

Time series classification has attracted considerable attention in both academia and industry \cite{ismail2019deep,liu2024generative}. Early approaches primarily relied on distance-based methods, such as Dynamic Time Warping (DTW) \cite{DTW} in combination with K-Nearest Neighbors (K-NN), which are effective in addressing temporal distortions in time series data. To overcome the limitations of these methods, ensemble techniques were introduced to improve classification accuracy. For instance, HIVE-COTE~\cite{cote} is an ensemble learning algorithm that enhances performance by combining multiple feature transformations and classifiers through hierarchical voting, offering a more comprehensive and robust representation of time series characteristics. With the advent of deep learning, more sophisticated models began to surpass traditional methods by automatically learning hierarchical features from raw data. Early deep learning architectures, such as fully convolutional networks~\cite{FCN} and recurrent neural networks~\cite{rnn}, demonstrated notable improvements in capturing local and sequential dependencies. More recently, models like InceptionTime~\cite{InceptionTime} have employed deeper networks with multi-scale convolutions, significantly improving the ability to recognize complex patterns across varying time scales. Additionally, Transformer-based models~\cite{informer,cheng2023formertime} have emerged as powerful alternatives, excelling at capturing long-range dependencies and global context, and further pushing the boundaries of time series classification performance across diverse application domains.

\subsection{LLMs in Time Series Analysis}
Given the impressive capabilities of LLMs, researchers are increasingly exploring their applications in time series analysis~\cite{gpt4ts,instruct}.
LLM-based approaches can be divided into two categories: fine-tuning and generative modeling.
Fine-tuning methods, such as Linear Fine-Tuning, combine pre-trained LLMs with time series-specific encoders, leveraging their linguistic capabilities to identify patterns. Generative models, like GPT-based forecasting, predict future time series sequences, while models like TEMPO integrate domain knowledge to enhance performance.

Despite effectiveness, LLMs in time series analysis encounter several significant challenges~\cite{weakness1,weakness2,tan2024language}. First, they struggle to capture temporal dependencies and channel-specific features, which are essential for accurate modeling. Second, a misalignment exists between numerical time series data and LLMs’ semantic space, which complicates processing. Third, fine-tuning these models is computationally expensive, particularly for large-scale applications. Lastly, LLMs fail to fully leverage their reasoning capabilities, limiting their potential. To address these issues, we propose TableTime, a paradigm for time series analysis through table understanding.

\section{Preliminaries}
\subsection{Problem Definitions}
Let $\mathbb{D} = \left\{ (X^1,y^1),(X^2,y^2),...,(X^n,y^n)\right\}$ be a dataset consisting of $n$ pairs $\left( X^i, y^i \right)$, where each $X^i \in \mathbb{R}^{t \times m}$ represents a multivariate time series with $t$ time steps and $m$ features, and $y^i \in \{ c_1, c_2, \dots, c_k \}$ is the corresponding label. The goal of the classification task is to learn a classifier on $\mathbb{D}$ that maps the input space $X$ to a probability distribution over the class $y$. In the context of TableTime, we propose using prompt engineering based on the characteristics of the dataset and task. Let $P$ denote the prompt, the model's output can be summarized as follows: $T^i = \text{LLM}(P, X^i)$ where $T^i$ is the text generated by the large language model in response to the prompt and the input time series $X^i$. To predict the label $y^i$ for each instance, we apply one regular expression to match $T^i$ and extract the classification result $\hat{y}^i$.

\subsection{Large Language Models}
Recent advancements in LLMs have unveiled a broad range of powerful capabilities, allowing for addressing a variety of complex tasks. In this context, we propose leveraging LLMs to enhance MTSC.
Several key advantages arise from utilizing LLMs to advance classification techniques:
\begin{itemize}[leftmargin=*]
\item \textit{World Knowledge}: Pre-trained on vast amounts of textual data from various domains, LLMs can integrate general knowledge into generations, enabling LLMs to provide contextual insights.

\item \textit{Reasoning}: LLMs exhibit advanced reasoning and pattern recognition abilities, which can potentially improve classification accuracy by capturing higher-level concepts.

\item \textit{Training-Free Inference}: LLMs have demonstrated remarkable training-free inference capabilities, showcasing their potential to generalize across domains without task-specific retraining. 

\item \textit{Text Generation}: LLMs solve the problem in the paradigm of text generation, which provides more possibilities for enhancing the interpretability of classification results.
\end{itemize}

\section{The Proposed TableTime}
\subsection{Model Architecture Overview}
An overview of the TableTime is shown in Figure~\ref{main_figure}. 
TableTime introduces an innovative approach to multivariate time series classification (MTSC) by leveraging large language models (LLMs). 
To enhance reasoning capabilities, we first employ neighbor retrieval to identify relevant neighbors, improving the understanding of LLMs. 
These raw numerical time series are then converted into tabular format, preserving both temporal and channel information.
To guide the LLM’s reasoning process, we design a comprehensive prompt that includes domain context information, neighbor knowledge, and task decomposition.
In the final step, TableTime applies reasoning capability to classify the test sample.

\subsection{Context Information Modeling}
\subsubsection{Reformulating Time Series as Tabular Data}
LLM-based models either learn time series embeddings directly in their latent space or align outputs from external models, often resulting in significant information loss, including temporal dependencies and channel relationships.
The inherent mismatch between numerical time series and textual semantic spaces introduces inefficiencies and limits the LLMs' ability to capture complex patterns in time series data.

To address these challenges, we propose table encoding, which reformulates time series as tabular format.
Table serves as a natural representation of time series, allowing for the preservation of both temporal and channel-specific information.
Through table encoding, the original multivariate time series are converted into a structured tabular form, which is then serialized into text for further processing. This process can be formalized as follows:
\begin{equation}
         X' = 
\begin{pmatrix}
0 & \mathbf{C}^{\top} \\
\mathbf{T} & X
\end{pmatrix} = 
\begin{pmatrix}
0 & c_1 & c_2 & \cdots & c_m \\
t_1 & x_{1,1} & x_{1,2} & \cdots & x_{1,m} \\
t_2 & x_{2,1} & x_{2,2} & \cdots & x_{2,m} \\
\vdots & \vdots & \vdots & \ddots & \vdots \\
t_t & x_{t,1} & x_{t,2} & \cdots & x_{t,m}
\end{pmatrix} ,
\end{equation}
where $\mathbf{C}^{\top} = \begin{pmatrix} c_1 & c_2 & \dots & c_m \end{pmatrix}$ denotes the channel-specific information and
$\mathbf{T} = \begin{pmatrix} t_1 & t_2 & \dots & t_t \end{pmatrix}^{T}$ denotes the timestamps. 

Subsequently, we convert the tabular time series data into a serialized textual format suitable for input. The process is formalized as follows: $\text{Text} = \text{Serialize}(X')$, where \text{Text} represents the serialized text, and \text{Serialize($\cdot$)} refers to the serialization function, such as DFLoader or MarkDown~\cite{encoding}.
Reformulating time series data into tabular format provides key advantages by preserving temporal dependencies and representing each channel separately, thereby enhancing both interpretability and compatibility.
In this format, each row corresponds to a timestamp, and each column represents a channel, enabling independent feature processing while maintaining sequential relationships.

\subsubsection{Domain Context Information}
While LLMs acquire extensive task-specific knowledge during pretraining, they remain sensitive to prompt design. Without explicit and well-structured instructions, they may misinterpret task intent, generate irrelevant content, or deviate from expected behavior. To address this, we incorporate domain context into the prompt, embedding key domain information to guide reasoning more effectively.

The domain context follows a structured template to ensure clarity, consistency, and semantic alignment with the task. It contains three components: (1) task definition — a concise description of the task within its domain; (2) dataset description — details of the dataset’s structure, length, and properties; and (3) class description — definitions and scopes of each label. This structured approach reduces ambiguity, aligns model reasoning with task constraints, and improves output consistency and reliability.

\subsection{Neighbor-Assisted In-Context Reasoner}
A major challenge in training-free MTSC with LLMs is the lack of prior exposure to test samples, leading to semantic ambiguity and limiting reasoning over complex temporal patterns. To address this, we propose a neighbor retrieval–augmented framework, termed neighbor-assisted in-context reasoning, which retrieves semantically relevant training examples and embeds them in the prompt to provide inductive cues. These include positive samples with similar dynamics and negative samples with contrasting behavior, helping the model anchor its predictions. We employ two retrieval strategies: positive sample guidance and contrast enhancement.

\subsubsection{Positive Sample Guidance}
To retrieve positive samples, we apply the k-nearest neighbors (KNN) algorithm to identify the set of \( k \) closest samples from the training dataset \( \mathbb{D}^{train} \) for each test sample \( X^{test} \). Formally, the positive sample set is defined as:
\begin{equation}
\mathcal{N}(X^{test}) = \operatorname{TopK} \left( \mathbb{D}^{train},\ F_{dist}(X^{test}, X^i) \right),
\end{equation}
where \( \operatorname{TopK}(\cdot) \) selects the \( k \) samples with the nearest distances to \( X^{test} \), \( F_{dist} \) is a chosen distance metric such as Euclidean distance, and \( X^i \in \mathbb{D}^{train} \) denotes the \( i \)-th training sample.

\subsubsection{Contrast Enhancement}
To further assist LLMs in classification, we introduce contrast enhancement mechanism that incorporates negative samples into the model’s reasoning process.
We begin by clustering the training dataset $\mathbb{D}^{train}$ using K-means to obtain a set of clusters \( \{C_k\}_{k=1}^K \), which is defined by the following optimization problem:
\begin{equation}
\{C_k\}_{k=1}^K = \mathop{\arg\min}\limits_{\{C_k\}_{k=1}^K}  \sum_{k=1}^K \sum_{x_i \in C_k} \|X_i - \mu_k\|^2,
\end{equation}
where \( \mu_k = \frac{1}{|C_k|} \sum_{X_i \in C_k} X_i \) denotes the centroid of cluster \( C_k \).

Given a test sample \( X^{test} \), we assign it to the cluster whose centroid is nearest:
\begin{equation}
j = \mathop{\arg\min}_{k} \| X^{test} - \mu_k \|,
\end{equation}

To construct negative samples for \( X^{test} \), we select from all clusters except the \( j \)-th cluster:
\begin{equation}
\mathcal{N}(X^{test}) = \operatorname{TopM} \left( \bigcup_{k \neq j} C_k, \ F_{dist}(X^{test}, X^i) \right),
\end{equation}
where \( \operatorname{TopM}(\cdot) \) denotes selecting the top \( M \) samples with the largest distances to \( X^{test} \), measured by a chosen distance metric (e.g., Euclidean distance) and \( X^i \in \mathbb{D}^{train} \) denotes the \( i \)-th training sample.
This ensures that the negative samples are sufficiently dissimilar from the test sample, enhancing contrastive learning.

\begin{figure}
    \centering
    \includegraphics[width=0.48\textwidth]{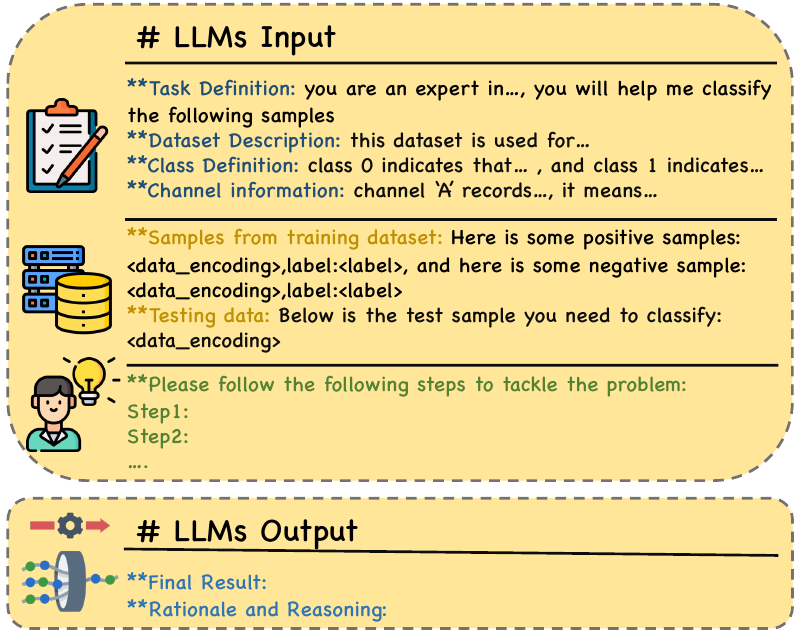}
    \caption {Prompt template of TableTime. There are three components of input from top to bottom.}
    \Description{template of TableTime. There are three components of input from top to bottom.}
    \label{Prompt}
\end{figure}

\subsection{Task Decomposition Mechanism}
Previous prompts often lack the structured guidance needed for LLMs to handle complex tasks effectively. This absence of clear instructions forces LLMs to interpret the task independently for each sample, which may result in inconsistencies or irrelevant reasoning.
Recently, step-by-step reasoning significantly improves the reasoning ability of LLMs~\cite{COT}. Based on this insight, we introduce task decomposition, which breaks MTSC into a series of smaller and manageable steps, allowing for more effective reasoning.

Instead of manually crafting the task decomposition strategy, we leverage a separate~\textit{Planning LLM} to generate it. Given a single labeled MTSC example, the Planning LLM is prompted to explain the classification process step by step. This self-derived reasoning path is then abstracted into a general decomposition template. During inference, a different~\textit{Reasoning LLM} which has no access to labels uses this structured prompt to perform classification. This separation ensures that the reasoning process aligns with the model's internal logic while preserving training-free conditions.

\subsection{Multi-Path Ensemble Enhancement}
Ensemble methods have demonstrated effectiveness for MTSC~\cite{minirocket,cote}.
When utilizing LLMs for MTSC, it cannot be ignored that the inherent stochasticity in LLM outputs makes direct classification challenging. In the meanwhile, self-consistency~\cite{wang2022self} leverages multiple outputs from one single LLM, selecting the most consistent response to enhance coherence and accuracy.

Building on the aforementioned considerations, we propose the multi-path ensemble enhancement strategy.
Specifically, we perform multiple inferences on the input $X^{test}$ under a set of distinct parameters of LLMs \( \{T_i\}_{i=1}^M \), where each parameter \( T_i \) induces a different reasoning trajectory within the LLM.
The classification from the \( i \)-th parameter is denoted as \( f_{T_i}(X^{test}) \).
The final result is obtained by aggregating the results via majority voting:
\begin{equation}
    y_{\mathrm{final}} = \arg\max_{y \in \mathcal{Y}} \sum_{i=1}^M \mathbb{I}\big(f_{T_i}(X^{test}) = y_i),
\end{equation}
where $y_i$ denotes the predicted labels of $f_{T_i}(\cdot)$
classifier and \(\mathbb{I}(\cdot)\) is the indicator function.
This approach leverages the diversity introduced by varying parameters to sample multiple plausible reasoning paths, thereby mitigating randomness from any single inference and improving robustness and accuracy on complex multivariate time series classification tasks.

\begin{table}[t]
	\centering
	\caption{Statistics of each dataset in the experiments.}
    \resizebox{0.48\textwidth}{!}{
	\begin{tabular}{ccccccc}
	\toprule
    Dataset & Train Size & Test Size & Dimensions & Length & Classes \\ 
    \midrule
    AWR     & 275        & 300       & 9          & 144    & 25       \\
    AF    & 15         & 15        & 2          & 640    & 3      \\
    BL      & 500        & 450       & 4          & 510    & 2    \\ 
    CR      & 108        & 72        & 6          & 1,197  & 12      \\
    ER      &30          &270        &4           & 65     &6  \\  
    FD    &5890 &3524 &144 &62 & 2\\  
    FM   & 316  & 100  & 28         & 50     & 2       \\
    SRS2 & 200 & 180       & 7          & 1,152  & 2       \\
    SWJ  & 12   &15         &4           &2500    &3 \\
    UWG  & 120        & 320       & 3          & 315    & 8       \\ 
		\bottomrule
	\end{tabular}
    }
	\label{UEA}
\end{table}

\subsection{Prompt Construction}
In TableTime, a structured prompt is crucial to achieve training-free classification.
As the template shown in Figure~\ref{Prompt}, the prompt includes three parts: (1) \textbf{domain context information} which provides LLMs with professional knowledge to warm up,
(2) \textbf{neighbor information} which provides crucial context by linking the test sample to similar or unsimilar labeled examples from the training dataset and 
(3) \textbf{task decomposition} that guides LLMs to implement step-by-step reasoning to generate the final results.

\subsection{Remark and Discussion}
In the following, we summarize the characteristics of TableTime and discuss its relations to Table, Dist and LLM-based models.

\begin{itemize}[leftmargin=*]
    \item \textit{Relation to Table-based Methods.}
    Unlike traditional tabular models relying on handcrafted features, TableTime uses tables as semantic prompts, enabling direct reasoning over temporal structures.
    \item \textit{Relation to Dist-based Methods.}
    While distance-based methods are interpretable but noise-sensitive, TableTime preserves interpretability and improves robustness via LLM-driven inference.
    \item \textit{Relation to LLM-based Methods.}
    Unlike costly LLM-based approaches with external embeddings, TableTime encodes raw time series into prompt-friendly tables, supporting training-free classification without representation loss.
\end{itemize}

\begin{table*}[t!]
  \centering
  \caption{Multivariate time series classification performance of TableTime and baseline models across ten datasets. The best are highlighted in bold, while the second-best are \underline{underlined}. "Best" indicates the frequency of achieving the highest accuracy.}
    \begin{tabular}{c|ccccccccccc|c}
    \toprule
    Model & AWR & AF    & BL    & CR    & ER    & FD    & FM    & SRS2  & SWJ   & UWG   & Average & Best \\
    \midrule
    nn-DTW & 0.9667  & 0.3333  & 0.5667  & 0.9654  & 0.9333  & 0.5184  & 0.5400  & 0.4889  & 0.2000  & 0.8906  & 0.6403  & 0  \\
    HIVE\_COTE V1 & 0.9756  & 0.2667  & \textbf{0.9978}  & 0.9547  & 0.9430  & 0.5376  & 0.4900  & 0.5222  & 0.6000  & 0.8875  & 0.7175  & 1  \\
    HIVE\_COTE V2 & 0.9233  & 0.2000  & \underline{0.9956}  & 0.9306  & 0.9222  & 0.5474  & 0.5300  & 0.5278  & 0.4000  & 0.8725  & 0.6849  & 0  \\
    MLP   & \underline{0.9822}  & 0.3556  & 0.7259  & \underline{0.9954}  & 0.7642  & 0.6690  & 0.5933  & 0.5611  & 0.4889  & 0.6281  & 0.6764  & 0  \\
    MiniRocket & 0.9433  & 0.4444  & 0.8793  & 0.9537  & \textbf{0.9728}  & 0.6065  & 0.5567  & 0.5370   & \underline{0.6444}  & \textbf{0.9365}  & 0.7475  & \underline{2} \\
    InceptionTime & 0.9807  & 0.4533  & 0.9165  & \textbf{0.9972}  & 0.8859  & \textbf{0.6818}  & 0.5985  & 0.5789  & 0.5733  & 0.8956  & \underline{0.7562}  & \underline{2} \\
    MCNN  & 0.9767  & 0.3778  & 0.9556  & 0.9259  & 0.9222  & 0.6747  & 0.5567  & 0.5648  & 0.5333  & 0.8563  & 0.7344  & 0  \\
    MCDCNN & 0.9789  & 0.4444  & 0.9763  & 0.9583  & 0.9358  & 0.5000  & 0.5800  & 0.5407  & 0.6000  & 0.8552  & 0.7370  & 0 \\
    TCN   & 0.9033  & 0.4000  & 0.7904  & 0.9537  & 0.5667  & 0.6801  & 0.5100  & 0.5148  & 0.3778  & 0.7531  & 0.6450  & 0  \\
    ConvTimeNet & \textbf{0.9844}  & 0.4444  & 0.9544  & 0.9769  & 0.8111  & 0.6613  & \underline{0.6100}  & \underline{0.5852}  & 0.3333  & 0.8635  & 0.7225  & 1  \\
    AutoFormer & 0.5733  & \underline{0.4667}  & 0.5881  & 0.7917  & 0.6296  & 0.5749  & 0.5600  & 0.5167  & 0.4444  & 0.4906  & 0.5636  & 0  \\
    Informer & 0.9811  & 0.2889  & 0.9378  & 0.9444  & 0.9432  & 0.5936  & 0.5867  & 0.5611  & 0.4667  & 0.8656  & 0.7169  & 0  \\
    TimesNet & 0.9800  & 0.3333  & 0.9474  & 0.9213  & 0.9185  & 0.6109  & 0.5700  & 0.5574  & 0.4222  & 0.8656  & 0.7127  & 0  \\
    GPT4TS & 0.9778  & 0.3778  & 0.9296  & 0.9352  & 0.9358  & 0.5821  & 0.5867  & 0.5611  & 0.4444  & 0.8542  & 0.7185  & 0  \\
    CrossTimeNet & 0.9367  & 0.3333  & 0.9521  & 0.9761  & 0.8296  & 0.5649  & 0.6058  & 0.5709  & 0.4667  & 0.7750  & 0.7011  & 0  \\
    Time-LLM & 0.7333  & 0.4000  & 0.5556  & 0.7083  & 0.7519  & 0.5482  & 0.6000  & 0.5667  & 0.6000  & 0.4438  & 0.5908  & 0  \\
    Time-FFM & 0.9733  & 0.3333  & 0.6911  & 0.9415  & 0.8667  & 0.5963  & 0.5600  & 0.5722  & 0.4000  & 0.8688  & 0.6803  & 0  \\
    \midrule
    TableTime & 0.9733  & \textbf{0.6667}  & 0.9222  & 0.9822  & \underline{0.9518}  & \underline{0.6771}  & \textbf{0.6400}  & \textbf{0.5889}  & \textbf{0.7333}  & \underline{0.8906}  & \textbf{0.8026}  & \textbf{5}  \\
    \bottomrule
    \end{tabular}
  \label{main}
\end{table*}

\begin{table*}[!h]
    \centering
    \caption{Experimental results of ablation results for five key modules. Evaluated by accuracy.}
    
    \begin{tabular}{l|ccccccc}
        \toprule
        Variants & AWR & AF & CR & FM & SRS2 & Average & IMP(\%) \\
        \midrule
        TableTime & \textbf{0.9733} & \textbf{0.6667} & \textbf{0.9822} & \textbf{0.6400} &\textbf{0.5889} & \textbf{0.7627} & -- \\
        \midrule
        \textit{w/o} timestamps & 0.9700 & 0.5333 & 0.9583 & 0.5500 & 0.5388 & 0.7101 & -7.81 \\
        \textit{w/o} channel information & 0.9633 & 0.5333 & 0.9444 & 0.5700 & 0.5500 & 0.7122 & -7.53 \\
        \textit{w/o} timestamps \& channel information& 0.9633 & 0.5333 & 0.9444 & 0.4200 & 0.5000 & 0.6722 & -12.73 \\
        \textit{w/o} domain context information & 0.9700 & 0.4000 & 0.9722 & 0.6000 & 0.5278 & 0.6940 & -9.90 \\
        \midrule
        \textit{w/o} negative samples & 0.9433 & 0.3867 & 0.9213 & 0.5334 & 0.5114 & 0.6592 & -14.41 \\        
        \bottomrule
    \end{tabular}
    \label{ablation}
\end{table*}

\section{Experiments}
\subsection{Experimental Setup}
\subsubsection{Datasets}
We conduct experiments on ten representative datasets from the well-known UEA MTSC archive~\cite{UEA}. The UEA archive has become one of the most widely used multivariate time series benchmarks.
Due to computational constraints, we use a set of 10 multivariate datasets from the UEA archive, which exhibit diverse characteristics in terms of the number and length of time series samples, as well as the number of categories. 
Specifically, we use the following datasets: ArticularyWordRecognition (AWR), AtrialFibrillation~(AF), Blink~(BL), Cricket~(CR), Ering~(ER),
FaceDetection~(FD), FingerMovements~(FM), StandWalkJump~(SWJ), SelfRegulationSCP2~(SRS2), , UWaveGestureLibrary~(UWG). In these original dataset, training and testing set have been well processed. We do not take any processing for a fair comparison. We summarize the main characteristics of dataset in Table \ref{UEA}.

\subsubsection{Baselines}
To conduct a comprehensive and fair comparison, we use baseline methods which is highly related to TableTime.
\textbf{Distance-based} methods: nn-DTW~\cite{DTW}.
\textbf{Hybrid} methods: HIVE-COTE V1~\cite{lines2018time}, HIVE-COTE V2~\cite{middlehurst2021hive}.
\textbf{Deep learning-based} methods: MLP~\cite{ismail2019deep}, MiniRocket~\cite{minirocket}, InceptionTime~\cite{InceptionTime}, MCNN~\cite{MCNN}, MCDCNN~\cite{MCDCNN}, TCN~\cite{TCN}, ConvTimeNet~\cite{convtimenet}, AutoFormer~\cite{autoformer}, Informer~\cite{informer} and TimesNet~\cite{timesnet}.
\textbf{LLM-based} methods: GPT4TS~\cite{gpt4ts}, CrossTimeNet~\cite{crosstimenet}, Time-LLM~\cite{timellm} and Time-FFM~\cite{timeffm}.

\subsubsection{Implement Details}
For nn-DTW and HIVE-COTE V1 and V2, we adopt the code in \footnote{\url{https://github.com/aeon-toolkit/aeon}}.
For the MCDCNN, MCNN, MiniRocket, MLP, TCN and InceptionTime, we adopt the publicly available code in \footnote{\url{https://github.com/timeseriesAI/tsai}}. 
For other baseline methods, we directly adopt the official implementation code. For a fair comparison, all models are trained on the training set and report the accuracy score on the testing set.
We use DTW as the neighbor retrieval methods. We use GPT-4o to generate the task decomposition and we use Llama-3.1-70b-instruct to tackle inference task. Temperature, top-p, max-tokens is setting to 0.2, 0.7 and 4096. Each experiment is repeated three times to minimize the uncertainty of the generated content of LLMs. All experiments in this paper set the number of inference paths to 3, and different inference paths come from different temperature settings, which are 0.1, 0.2, and 0.3.

\begin{figure*}[h]
    \centering
    \includegraphics[width=\textwidth]{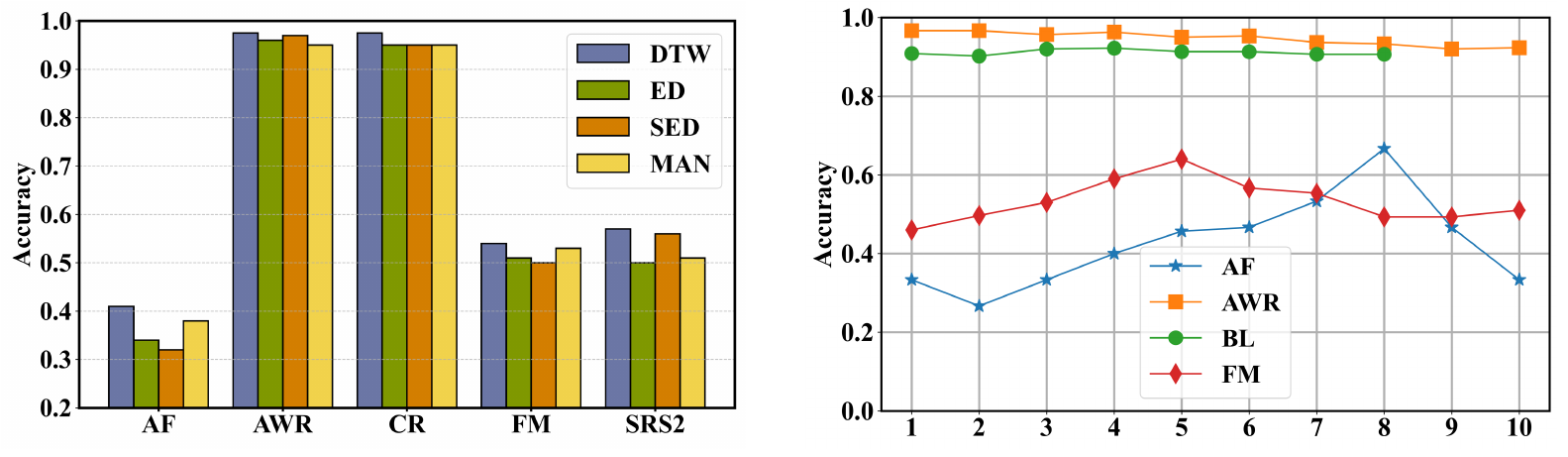}
    \caption{(\textbf{Left}) : Experimental results of four different neighbor retrieval methods: DTW, ED, SED, MAN. (\textbf{Right}) : Classification accuracy results under different neighbor number settings.}
    \label{retrieval}
    \Description{(\textbf{Left}) : Experimental results of four different neighbor retrieval methods: DTW, ED, SED, MAN. (\textbf{Right}) : Classification accuracy results under different neighbor number settings.}
\end{figure*}

\subsection{Classification Results Analysis}
\begin{figure}
    \centering
    \includegraphics[width=0.5\textwidth]{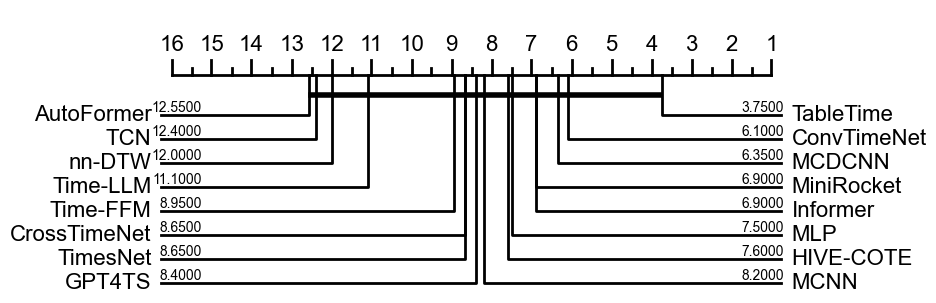}
    \caption {Critical difference diagram over the mean ranks of TableTime, baseline models.}
    \label{cd}
    \Description{Critical difference diagram over the mean ranks of TableTime, baseline models.}
\end{figure}

Table~\ref{main} summarizes the classification accuracy of all compared methods, and Figure~\ref{cd} reports the critical difference diagram as presented in \cite{critical}. Compared to other baseline models, the experimental results demonstrate that our proposed TableTime achieves competitive performance and notable advantages on several datasets. For each dataset, the performance of TableTime is either the most accurate one or very close to the best one. This all demonstrates the powerful capabilities of TableTime in MTSC tasks.

We observe that TableTime surpasses other baselines by a large margin on datasets like AF and SWJ, where both training and testing sets are small, highlighting the strong training-free classification ability of LLMs in data-scarce scenarios. Our method also consistently outperforms GPT4TS, Time-LLM, and Time-FFM, confirming its effective use of LLM reasoning for superior classification. However, TableTime lags behind optimal methods on BL and UWG, likely because their large scale enables deep learning models to capture richer features.

\subsection{Study of Context Information Modeling}
\subsubsection{Effectiveness of Timestamps and Channel Information}
To assess the importance of timestamps and channel information, we conduct three ablation experiments: ablation of timestamps, ablation of channel information, and ablation of both.
As shown in Table~\ref{ablation}, it is evident that both timestamps and channel information play a crucial role in the performance of TableTime.
Among these, the former is more important, which further confirms the importance of temporal information for time series modeling.

\subsubsection{Effectiveness of Domain Context Information}
To assess the effectiveness of domain context information in TableTime, we conduct its ablation study.
As shown in Table~\ref{ablation}, we reveal the importance of it in assisting LLM reasoning, which is an integral part for TableTime to make inferences that conform to domain rules.

\subsubsection{Various Table Formats}
Table format influences performance in two ways: (1) different encodings yield different token counts, and (2) they lead to varied LLM outputs due to differing levels of structural information. We evaluate several encoding methods by computing the mean accuracy of all neighbors under each format (Table~\ref{encoding}). DFLoader performs best, likely because it represents each channel separately, avoiding feature mixing. In contrast, JSON encodes values per time step, limiting comprehension; HTML introduces many irrelevant characters that may hinder reasoning; and MarkDown blurs the distinction between time indices and feature channels, affecting interpretation.

\begin{table}[t]
\centering
\caption{Comparison of four table formats on four datasets. DFLoader contributes the best encoding method.}
\begin{tabular}{ccccc}
\hline
Table Format & AWR & AF &FM &UWG \\
\hline
DFLoader   & \textbf{0.9733} & \textbf{0.6667} & \textbf{0.6400} & \textbf{0.8906} \\
HTML       & 0.9233 & 0.4867  &0.5621 &0.8241\\
JSON       & 0.9567 & 0.5433  &0.5786 &0.8564\\
MarkDown   & 0.9654 & 0.5574  &0.5984 &0.8746\\
\hline
\end{tabular}
\label{encoding}
\end{table}

\subsection{Analysis of Neighbor-Assisted Strategy}
\subsubsection{Analysis of Neighbor Retrieval Methods}
Although no distance function achieves SOTA on all datasets, our analysis (Figure~\ref{retrieval}) shows that retrieval choice strongly impacts TableTime’s performance. Dynamic Time Warping~(DTW) outperforms Manhattan~(MAN), Euclidean~(ED), and Standardized Euclidean~(SED), likely due to its robustness in high-dimensional time series by flexibly aligning sequences to handle temporal distortions and shifts. The advantage is most pronounced on the AF dataset, while the gap narrows on larger datasets, suggesting smaller datasets are more sensitive to retrieval choice.

\begin{figure*}[!ht]
    \centering
    \includegraphics[width=\textwidth]{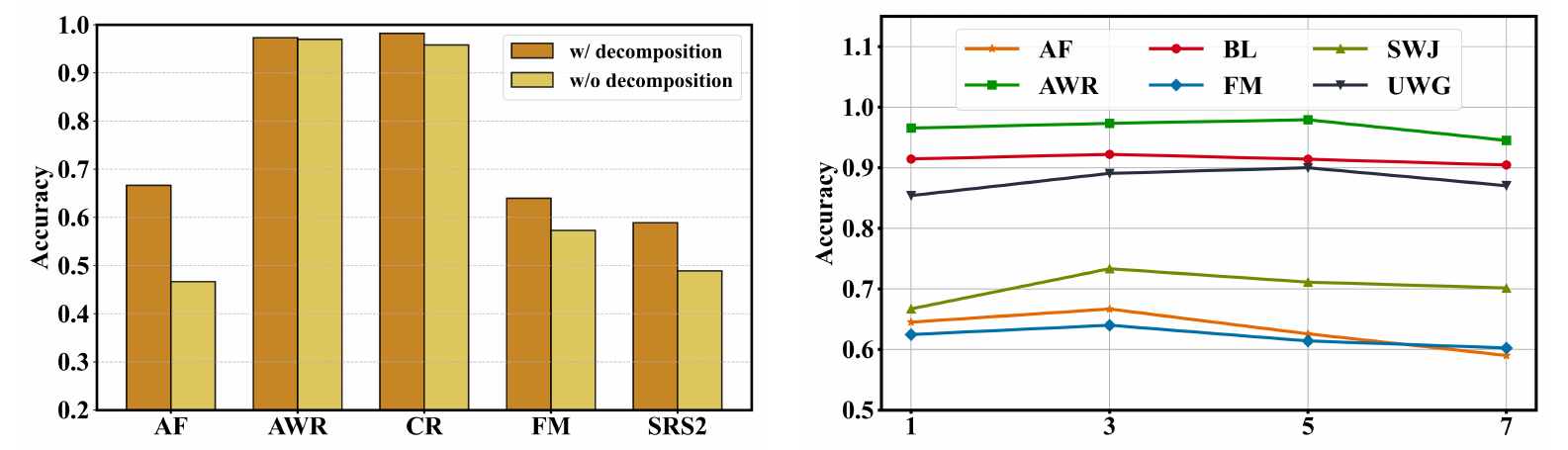}
    \caption{(\textbf{Left}) : Effectiveness of task decomposition mechanism in TableTime framework. (\textbf{Right}) : TableTime classification performance when the number of inference paths is 1, 3, 5, and 7.}
    \Description{(\textbf{Left}) : Effectiveness of task decomposition mechanism in TableTime framework. (\textbf{Right}) : TableTime classification performance when the number of inference paths is 1, 3, 5, and 7.}
    \label{reasoning}
\end{figure*}

\subsubsection{Study of the Number of Positive Samples}
In order to further explore the impact of the number of nearest neighbors on the final classification accuracy, we count the classification accuracy under different neighbor samples.
As shown in Figure~\ref{retrieval}, the accuracy initially increases as the number of nearest neighbors (K) increases, and it begins to decline beyond a certain point.
This pattern suggests that while a moderate increase in neighbors can enhance performance by providing relevant context, too many neighbors may introduce noise, leading to decreased accuracy. We attribute this decline to potential "model hallucination," where excessive contextual information makes it challenging for the model to filter out irrelevant data, thus reducing classification accuracy.

\begin{figure*}[!ht]
    \centering
    \begin{subfigure}[b]{0.45\textwidth}
        \includegraphics[width=\textwidth]{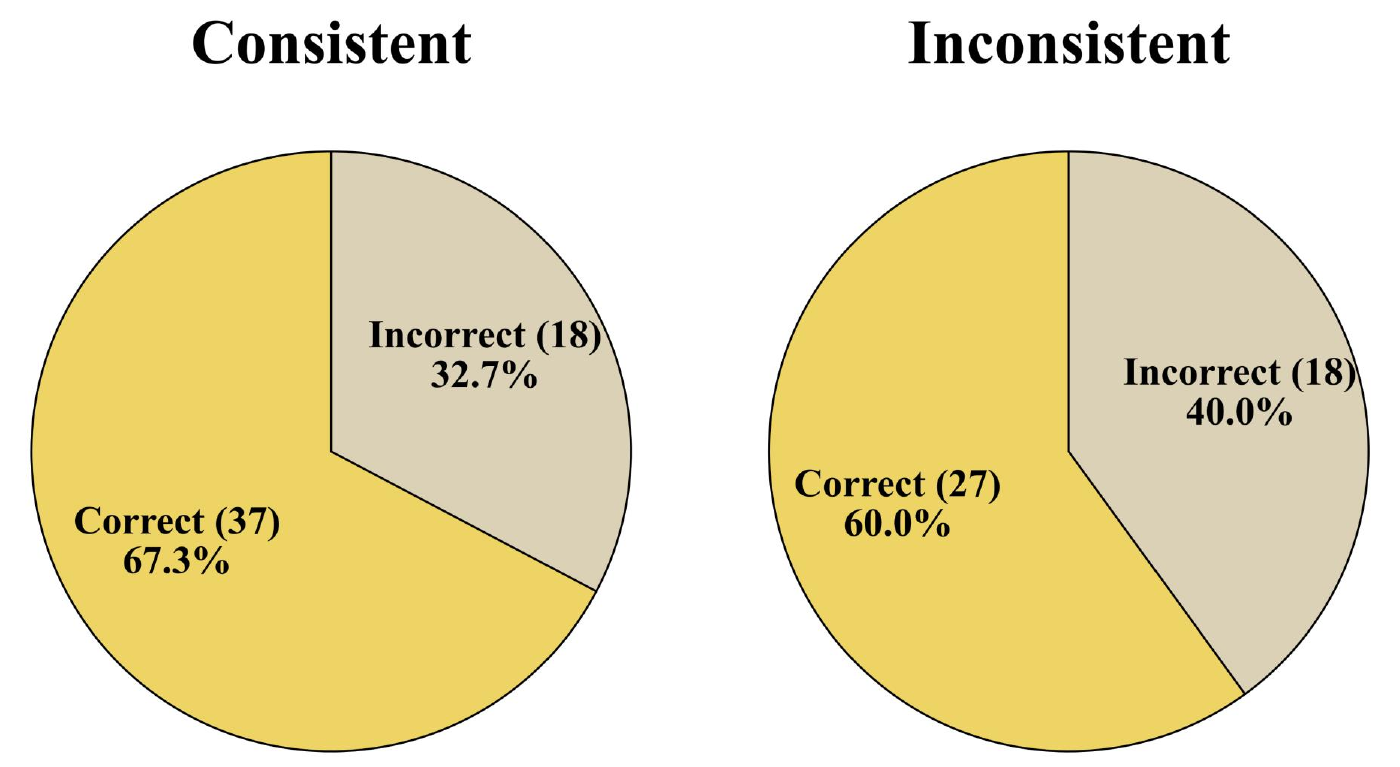}
        \caption{FM dataset.}
    \end{subfigure}
    \hfill
    \begin{subfigure}[b]{0.45\textwidth}
        \includegraphics[width=\textwidth]{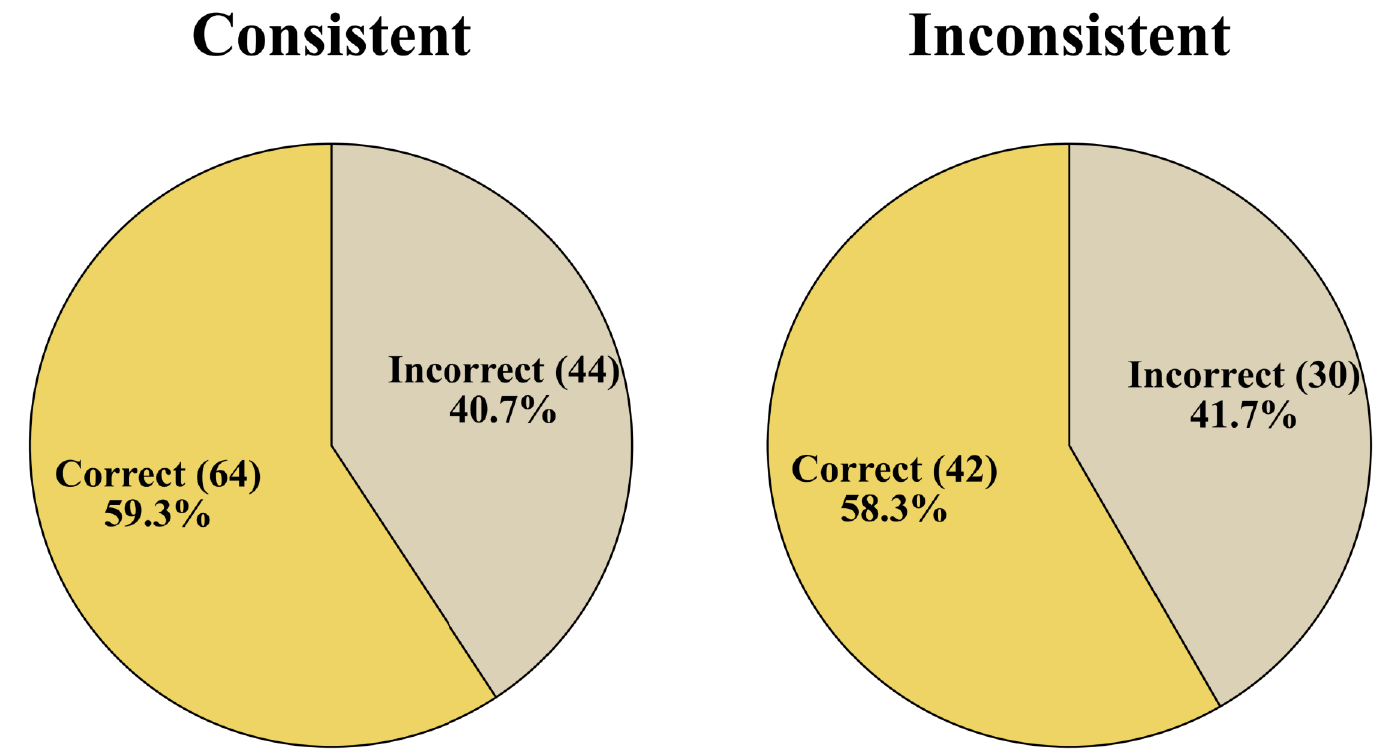} 
        \caption{SRS2 dataset.}
    \end{subfigure}
    \caption{Result of nearest neighbor consistency analysis.
The results show the classification results when the nearest
neighbor is selected and not selected.}
\Description{Result of nearest neighbor consistency analysis.
The results show the classification results when the nearest
neighbor is selected and not selected.}
    \label{consistency}
\end{figure*}

\subsubsection{Study of Negative Samples}
By retrieving negative samples, we can provide negative examples to the LLM and guide it to perform comprehensive reasoning. In this section, we verify whether negative samples are really effective for classifying TableTime. Specifically, we remove these negative samples in the prompt and only keep the positive samples in the prompt.

As the results shown in the Table~\ref{ablation}, the removal of negative samples results in a 14.41\% drop in classification performance. This outweighs the performance degradation caused by removing timestamps and channel information. This proves the irreplaceable role of negative samples in TableTime classification, which can help LLMs achieve deeper feature extraction during classification.

\subsection{Analysis of Reasoning Strategy}
\subsubsection{Effectiveness of Task Decomposition Mechanism}
Task decomposition systematically breaks down MTSC task into smaller and manageable steps, ensuring that the LLMs could effectively process the input timeseries and evaluate the temporal and channel-specific features
for decision making. To evaluate the impact of task decomposition, we conduct ablation experiments by removing this component from the prompt.
As shown in Figure~\ref{reasoning}, results indicate that removing this mechanism leads to noticeable drops in accuracy, particularly on datasets with high-dimensional and complex temporal patterns. The structured reasoning enabled by problem decomposition not only enhances interpretability but also strengthens the model’s ability to generalize, demonstrating its critical role in achieving robust and reliable performance.

\subsubsection{Analysis of Multi-path Ensemble Module}
The multi-path ensemble module aims to enhance classification robustness and accuracy by aggregating predictions from diverse inference paths generated through varying the temperature of the LLM. In this part, we explore the classification results for different number of temperature combinations.
As shown in Figure~\ref{reasoning}, multi-path ensemble reasoning improves classification performance over a single path, demonstrating its effectiveness. However, performance gains do not scale linearly with more paths. While increasing the number of inference paths from 1 to 3 and 5 improves accuracy on most datasets, using 7 paths leads to a performance drop and higher inference cost. This indicates a trade-off between performance and computational efficiency.

\subsection{In-Depth Evaluation w.r.t the Performance of TableTime}
This study examines the relationship between LLM predictions and the labels of their nearest neighbors, shedding light on the model’s dependence on neighbor information and its robustness when discrepancies occur. We categorize results into cases where predictions match or differ from neighbor labels and analyze correctness within each group. 
As shown in Figure~\ref{consistency}, TableTime maintains strong performance: even with inconsistent neighbors, accuracy remains above 50\% (60.0\% for FM, 58.3\% for SRS2). These results suggest that while consistent neighbors improve accuracy, the model can still rely on LLM reasoning to deliver robust predictions despite noisy or incorrect references.

\begin{figure}[!t]
    \centering
    \includegraphics[width=0.45\textwidth]{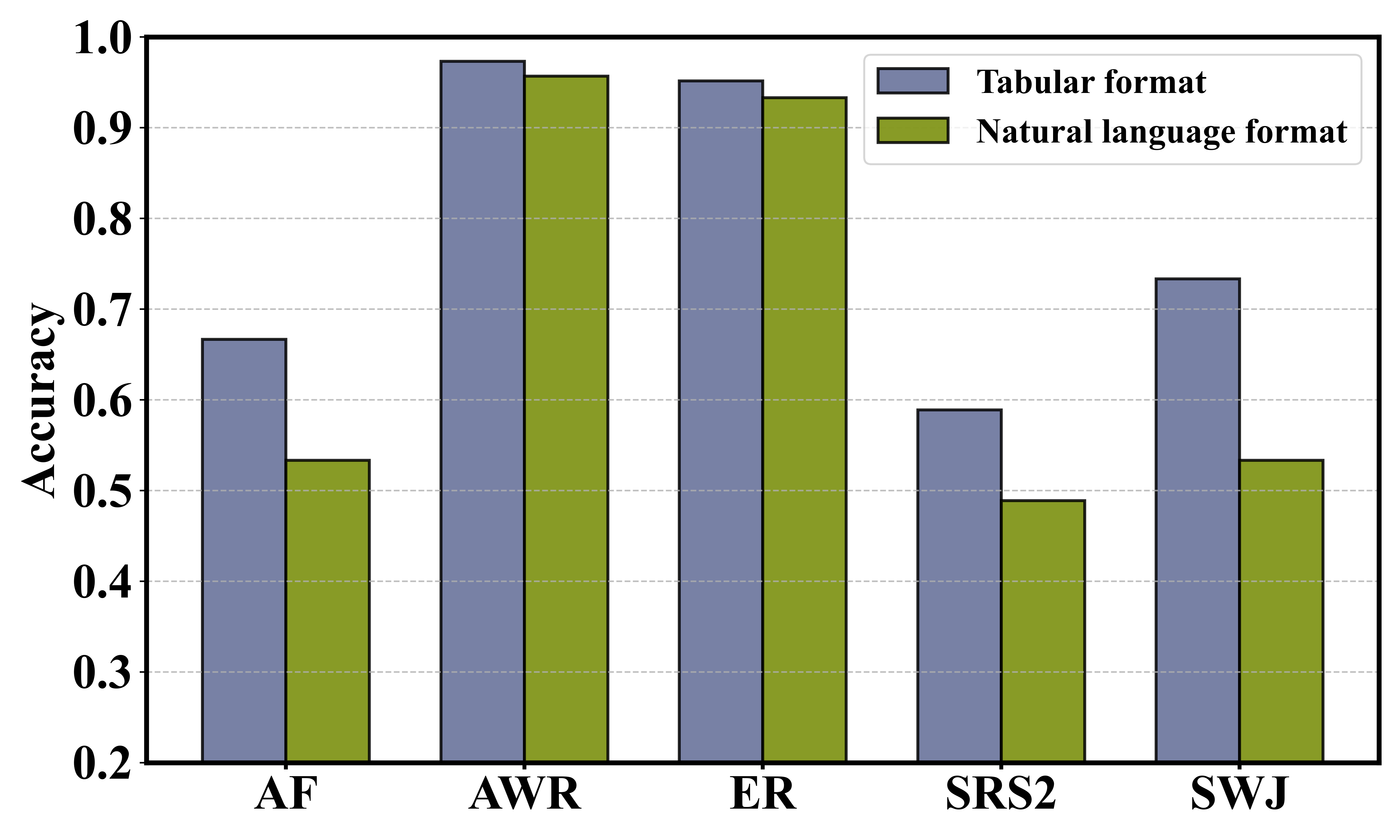}
    \caption{Comparison of classification performance of TableTime under tabular and natural language input.}
    \Description{Comparison of classification performance of TableTime under tabular and natural language input.}
    \label{form}
\end{figure}

\subsection{Tabular vs. Natural Language Inputs}
We investigate why tabular inputs are preferred over natural language. Although both formats encode the same data, tabular prompts explicitly align variables and values, whereas natural language embeds numbers in prose. Experiments on five datasets, replacing tabular input with sentences like “The channel is ... and its value is ...”, show that tabular formatting consistently achieves higher accuracy (Figure~\ref{form}). This indicates that structural alignment, rather than semantics alone, facilitates cross-variable and temporal reasoning, while prose descriptions add unnecessary parsing burden.

\subsection{Case Study}
\begin{figure}[ht]
    \centering
    \includegraphics[width=0.5\textwidth]{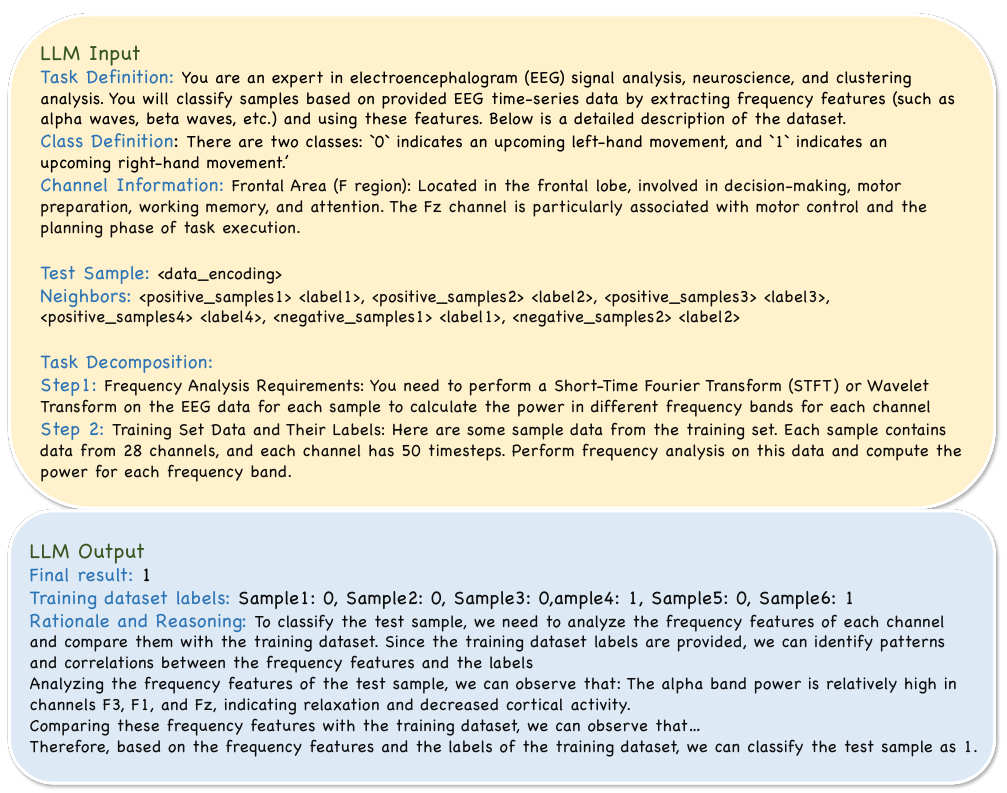}
    \caption{A case study of how TableTime tackles time series classification. This example shows how TableTime utilizing task decomposition to this task.}
    \Description{A case study of how TableTime tackles time series classification. This example shows how TableTime utilizing task decomposition to this task.}
    \label{case}
\end{figure}

Figure~\ref{case} shows one example of how TableTime tackles time series classification task. From this we can clearly see that TableTime's classification selection is not an ordinary clustering, but reflects rigorous multi-step reasoning thinking. In this example, we input four positive samples and two negative samples. Although the first three positive samples are opposite to the correct answer, TableTime can still give the correct classification result. This further proves the effectiveness of TableTime. At the same time, we can see that the input results of TableTime are very logical, which further confirms the effectiveness of the task decomposition mechanism.
\section{Conclusion and Limitation}
In this work, we highlight the critical importance of explicitly modeling temporal and channel-specific information in raw time series data. By converting time series into tabular time series, we naturally preserve the two information. Then we designed a reasoning-enhanced prompt to stimulate the reasoning ability of LLMs to training-free classification. From this perspective, we naturally reformulate the multivariate time series classification(MTSC) problem as a table understanding problem, providing a new paradigm for MTSC. We propose the TableTime, a training-free time series classification reasoning framework. The classification results on 10 datasets demonstrate the superior performance of our method and the possibility to become a new paradigm in the field of MTSC.

Despite the strengths of our model, we acknowledge several limitations that warrant further investigation. First, it is important to explore efficient methods for encoding tabular time series within our framework. As discussed earlier, certain encoding techniques may hinder the interpretability of LLMs. Second, the nearest neighbor retrieval process presents opportunities for optimization. Beyond performing retrieval directly on the original time series data, an alternative approach involves embedding the original data first and then conducting nearest neighbor retrieval. This method allows for a more comprehensive exploration of the features, enabling deeper insights.

\section*{Acknowledgments}
This research was supported by grants from the National Natural Science Foundation of China (No. 62502486), the grants of Provincial Natural Science Foundation of Anhui Province (No. 2408085QF193), the Fundamental Research Funds for the Central Universities of China (No. WK2150110032).

\newpage

\appendix

\section*{GenAI Usage Disclosure}
In the preparation of this manuscript, the authors employed generative AI tools solely for language editing purposes, such as correcting grammar, spelling, and improving sentence clarity. No generative AI technologies were used in the conception, experimental design, data analysis, or content generation of this paper. All intellectual contributions remain the sole work of the human authors. According to the ACM Policy on Authorship, such usage does not require disclosure; however, we include this statement for transparency.

\bibliographystyle{ACM-Reference-Format}
\balance
\bibliography{cite}
\clearpage
\end{document}